\documentclass[11pt]{article}
\usepackage{acl}

\usepackage{times}
\usepackage{graphics}
\usepackage{latexsym}
\usepackage{amsmath}
\usepackage{natbib}
\usepackage{hyperref}
\usepackage{xcolor}
\usepackage{enumitem}
\usepackage{graphicx}

\newcommand{\norm}[1]{\left\lVert#1\right\rVert}

\usepackage{microtype}

\title{BitFit: Simple Parameter-efficient Fine-tuning\\ for  Transformer-based Masked Language-models}

\author{Elad Ben-Zaken\textsuperscript{1} \, Shauli Ravfogel\textsuperscript{1,2} \,  \, Yoav Goldberg\textsuperscript{1,2}\\
\textsuperscript{1}Computer Science Department, Bar Ilan University \\
\textsuperscript{2}Allen Institute for Artificial Intelligence \\
  {\tt  \{benzakenelad, shauli.ravfogel, yoav.goldberg}{\}@gmail.com}
  }

\date{}

\begin{document}
\maketitle

\begin{abstract}
We introduce BitFit, a sparse-finetuning method where only the bias-terms of the model (or a subset of them) are being modified. We show that with small-to-medium training data, applying BitFit on pre-trained BERT models is competitive with (and sometimes better than) fine-tuning the entire model. For larger data, the method is competitive with other sparse fine-tuning methods.
Besides their practical utility, these findings are relevant for the question of understanding the commonly-used process of finetuning: they support the hypothesis that finetuning is mainly about exposing knowledge induced by language-modeling training, rather than learning new task-specific linguistic knowledge. 
\end{abstract}

\section{Introduction}

Large pre-trained transformer based language models, and in particular bidirectional masked language models from the BERT family \cite{BERT,ROBERTA,DBLP:journals/corr/abs-1907-10529}, are responsible for significant gains in many NLP tasks. Under the common paradigm, the model is pre-trained on large, annotated corpora with the LM objective, and then \emph{finetuned} on task-specific supervised data. 
The large size of these models make them expensive to train and, more importantly, expensive to deploy. This, along with theoretical questions on the extent to which finetuning must change the original model, has led researchers to consider fine-tuning variants where one identifies a small subset of the model parameters which need to be changed for good performance in end-tasks, while keeping all others intact (\S{\ref{sec:background}}). 

We present a simple and effective approach to fine tuning (\S\ref{sec:method}), which has the following benefits:
\begin{enumerate}[itemsep=-2px]
    \item Changing very few parameters per fine-tuned task.
    \item Changing the same set of parameters for every tasks (task-invariance).
    \item The changed parameters are both isolated and localized across the entire parameter space. 
    \item For small to medium training data, changing only these parameters reaches the same task accuracy as full fine-tuning, and sometimes even improves results.
\end{enumerate}
Specifically, we show that freezing most of the network and \textbf{fine-tuning only the bias-terms} is surprisingly effective. Moreover, if we allow the tasks to suffer a small degradation in performance, we can fine-tune only two bias components (the ``query" and ``middle-of-MLP" bias terms), amounting to half of the bias parameters in the model, and only 0.04\% of all model parameters.

This result has a large practical utility in deploying multi-task fine-tuned models in memory-constrained environments, as well as opens the way to trainable hardware implementations in which most of the parameters are fixed. Additionally, it opens up a set of research directions regarding the role of bias terms in pre-trained networks, and the dynamics of the fine-tuning process.

\section{Background: fine-tuning and parameter-efficient fine-tuning}
\label{sec:background}

In transfer-learning via model fine-tuning, a pre-trained encoder network takes the input and produces contextualized representations. Then, a task-specific classification layer (here we consider linear classifiers) is added on top of the encoder, and the entire network (encoder+task specific classifiers) is trained end-to-end to minimize the task loss.\\

\noindent\textbf{Desired properties.} While fine-tuning per-task is very effective, it also results in a unique, large model for each pre-trained task, making it hard to reason about what was changed in the fine-tuning process, as well as hard to deploy, especially as the number of tasks increases. Ideally, one would want a fine-tuning method that:\\
(i) matches the results of a fully fine-tuned model;\\
(ii) changes only a small portion of the model's parameters; and (iii) enables tasks to arrive in a stream, instead of requiring simultaneous access to all datasets. 
For efficient hardware based deployments, it is further preferred that (iv): the set of parameters that change values is consistent across different tasks.\\\\
\noindent\textbf{Learning vs. Exposing.} The feasibility of fulfilling the above requirements depends on a fundamental question regarding the nature of the fine-tuning process of large pre-trained LMs: to what extent does the fine-tuning process induces the \emph{learning of new capabilities}, vs. the \emph{exposing of existing capabilities}, which were learned during the pre-training process.\\\\
\noindent\textbf{Existing approaches.} Two recent works have demonstrated that adaptation to various end-tasks can in fact be achieved by changing only a small subset of parameters.\\
The first work, by \citet{DBLP:journals/corr/abs-1902-00751} (``Adapters''), achieves this goal by injecting small, trainable task-specific ``adapter" modules between the layers of the pre-trained model, where the original parameters are shared between tasks.\\
The second work, by \citet{diffp} (``Diff-Pruning''), achieves the same goal by adding a sparse, task-specific difference-vector to the original parameters, which remain fixed and are shared between tasks. The difference-vector is regularized to be sparse.\\
Both methods allow adding only a small number of trainable parameters per-task (criteria ii), and each task can be added without revisiting previous ones (criteria iii).\\
They also partially fulfill criteria (i), suffering only a small drop in performance compared to full fine-tuning.\\
The Adapter method, but not the Diff-Pruning method, also supports criteria (iv). However, Diff-Pruning is more parameter efficient than the Adapter method (in particular, it adds no new parameters), and also achieves better task scores.\\
We compare against Diff-Pruning and Adapters in the experiments section, and show that we perform favorably on many tasks while also satisfying criteria (iv).

\section{Bias-terms Fine-tuning (BitFit)}
\label{sec:method}

We propose a method we call BitFit\footnote{Our code is publicly available at \url{www.github.com/benzakenelad/BitFit}} (BIas-Term FIne-Tuning), in which we freeze most of the transformer-encoder parameters, and train only the bias-terms and the task-specific classification layer. BitFit has three key properties: (i) match the results of fully fine-tuned model. (ii) enable tasks to arrive in a stream, this way it does not require simultaneous access to all datasets. (iii) fine-tune only a small portion of the model's parameters.\par

The approach is parameter-efficient: each new task requires storing only the bias terms parameter vectors (which amount to less than 0.1\% of the total number of parameters), and the task-specific final linear classifier layer. 

Concretely, the BERT encoder is composed of $L$ layers, where each layer $\ell$ starts with $M$ self-attention heads, where a self attention head $(m,\ell)$ has \emph{key}, \emph{query} and \emph{value} encoders, each taking the form of a linear layer: 
\begin{align*}
\mathbf{Q}^{m,\ell}(\mathbf{x}) &=
 \textcolor{blue}{\mathbf{W}_{q}^{m,\ell}}\mathbf{x} + \textcolor{purple}{\mathbf{b}_{q}^{m,\ell}} \\
\mathbf{K}^{m,\ell}(\mathbf{x}) &= \textcolor{blue}{\mathbf{W}_{k}^{m,\ell}}\mathbf{x} + \textcolor{purple}{\mathbf{b}_{k}^{m,\ell}} \\
\mathbf{V}^{m,\ell}(\mathbf{x}) &= \textcolor{blue}{\mathbf{W}_{v}^{m,\ell}}\mathbf{x} + \textcolor{purple}{\mathbf{b}_{v}^{m,\ell}}
\end{align*}
Where $\mathbf{x}$ is the output of the former encoder layer (for the first encoder layer $\mathbf{x}$ is the output of the embedding layer).
These are then combined using an attention mechanism that does not involve new parameters:
\[ \mathbf{h}_1^{\ell} = att \big{(}\mathbf{Q}^{1,\ell},\mathbf{K}^{1,\ell},\mathbf{V}^{1,\ell},..,\mathbf{Q}^{m,\ell},\mathbf{K}^{m,\ell},\mathbf{V}^{m,l}\big{)} \]

\noindent and then fed to an MLP with layer-norm (LN):
\begin{align}
\mathbf{h}_{2}^{\ell} &= \text{Dropout}\big{(} \textcolor{blue}{\mathbf{W}_{m_1}^{\ell}} \cdot \mathbf{h}_{1}^{\ell}\;\;+\;\;\textcolor{purple}{\mathbf{b}_{m_1}^{\ell}} \big{)}\\
\mathbf{h}_{3}^{\ell} &= \textcolor{blue}{\mathbf{g}_{LN_1}^{\ell}}\odot\frac{(\mathbf{h}_{2}^{\ell} + \mathbf{x}) - \mu}{\sigma} + \textcolor{purple}{\mathbf{b}_{LN_1}^{\ell}}\\
\mathbf{h}_{4}^{\ell} &= \;\;\text{GELU}\big{(}\textcolor{blue}{\mathbf{W}_{m_2}^{\ell}} \cdot \mathbf{h}_{3}^{\ell} \;\;\;+\textcolor{purple}{\;\;\mathbf{b}_{m_2}^{\ell}}\big{)}\\
\mathbf{h}_{5}^{\ell} &= \text{Dropout}\big{(}\textcolor{blue}{\mathbf{W}_{m_3}^{\ell}} \cdot \mathbf{h}_{4}^{\ell}\;\;+\textcolor{purple}{\;\;\;\mathbf{b}_{m_3}^{\ell}}\big{)}\\
\mathbf{\;\;\;\text{out}}^{\ell} &= \textcolor{blue}{\mathbf{g}_{LN_2}^{\ell}}\odot\frac{(\mathbf{h}_{5}^{\ell} + \text{h}_{3}^{\ell}) - \mu}{\sigma} + \textcolor{purple}{\mathbf{b}_{LN_2}^{\ell}}
\end{align}

The collection of all matrices $\color{blue}\mathbf{W}_{(\cdot)}^{\ell, (\cdot)}$ and vectors $\color{blue}\mathbf{g}_{(\cdot)}^{\ell}$, $\color{purple}\mathbf{b}_{(\cdot)}^{\ell, (\cdot)}$, indicated in  \textcolor{blue}{blue} and \textcolor{purple}{purple} are the network's \emph{parameters} $\Theta$, where the subset of \textcolor{purple}{purple} vectors $\textcolor{purple}{\mathbf{b}_{(\cdot)}^{\ell, (\cdot)}}$ are the \emph{bias terms}\footnote{In Appendix \S\ref{app:names} we relate this notation with parameter names in HuggingFace implementation.}.

\begin{table*}[h!]
\centering
\scalebox{0.695}{
\begin{tabular}{llcccccccccccc}
\hline
 & & \textbf{\%Param} & \textbf{QNLI} & \textbf{SST-2} & \textbf{MNLI\textsubscript{m}} & \textbf{MNLI\textsubscript{mm}} & \textbf{CoLA} & \textbf{MRPC} & \textbf{STS-B} & \textbf{RTE} & \textbf{QQP} & \textbf{Avg.} \\
& Train size & & 105k & 67k & 393k & 393k & 8.5k & 3.7k & 7k & 2.5k & 364k & \\
\hline
(V) & Full-FT$\dagger$ & 100\% & \textbf{93.5} & \textbf{94.1} & \textbf{86.5} & \textbf{87.1} & \textbf{62.8} & \textbf{91.9} & 89.8 & 71.8 & \textbf{87.6} & \textbf{84.8}\\
(V) & Full-FT & 100\% & 91.7$\pm$0.1 & 93.4$\pm$0.2 & 85.5$\pm$0.4 & 85.7$\pm$0.4 & 62.2$\pm$1.2 & 90.7$\pm$0.3 & \textbf{90.0$\pm$0.4} & \textbf{71.9$\pm$1.3} & 87.5$\pm$0.4 & 84.1 \\
\hline
(V) & Diff-Prune$\dagger$ & 0.5\% & \textbf{93.4} & \textbf{94.2} & \textbf{86.4} & \textbf{86.9} & 63.5 & 91.3 & 89.5 & 71.5 & \textbf{86.6} & \textbf{84.6} \\
(V) & BitFit & 0.08\% & 91.4$\pm$2.4 & 93.2$\pm$0.4 & 84.4$\pm$0.2 & 84.8$\pm$0.1 & \textbf{63.6$\pm$0.7} & \textbf{91.7$\pm$0.5} & \textbf{90.3$\pm$0.1} & \textbf{73.2$\pm$3.7} & 85.4$\pm$0.1 & 84.2 \\
\hline
(T) & Full-FT$\ddagger$ & 100\%  & 91.1 & \textbf{94.9} & 86.7 & 85.9 & \textbf{60.5} & \textbf{89.3} & \textbf{87.6} & 70.1 & \textbf{72.1} & \textbf{81.8} \\
    (T) & Full-FT$\dagger$ & 100\% & \textbf{93.4} & 94.1 & 86.7 & \textbf{86.0} & 59.6 & 88.9 & 86.6 & \textbf{71.2} & 71.7 & 81.5 \\
\hline
(T) & Adapters$\ddagger$ & 3.6\% & 90.7 & 94.0 & 84.9 & 85.1 & 59.5 & 89.5 & \textbf{86.9} & 71.5 & \textbf{71.8} & 81.1 \\
(T) & Diff-Prune$\dagger$ & 0.5\% & \textbf{93.3} & 94.1 & \textbf{86.4} & \textbf{86.0} & \textbf{61.1} & \textbf{89.7} & 86.0 & 70.6 & 71.1 & \textbf{81.5} \\
(T) & BitFit & 0.08\% & 92.0 & \textbf{94.2} & 84.5 & 84.8 & 59.7 & 88.9 & 85.5 & \textbf{72.0} & 70.5 & 80.9 \\
\hline
\end{tabular}}
\caption{\label{table:main} BERT\textsubscript{LARGE} model performance on the GLUE benchmark validation set (V) and test set (T). Lines with $\dagger$ and $\ddagger$ indicate results taken from \citet{diffp} and \citet{DBLP:journals/corr/abs-1902-00751} (respectively).}
\end{table*}

The bias terms are additive, and correspond to a very small fraction of the network, in BERT\textsubscript{BASE} and BERT\textsubscript{LARGE} bias parameters make up 0.09\% and 0.08\% of the total number of parameters in each model, respectively.

We show that by freezing all the parameters $\color{blue}\mathbf{W}^{(\cdot)}$ and $\color{blue}\mathbf{g}^{(\cdot)}$ and fine-tuning only the additive bias terms $\color{purple}\mathbf{b}^{(\cdot)}$, we achieve transfer learning performance which is comparable (and sometimes better!) than fine-tuning of the entire network,

We also show that we can fine-tune only a subset of the bias parameters, namely those associated with the \emph{query} and the \emph{second MLP layer}  (only $\color{purple}\mathbf{b}_{q}^{(\cdot)}$ and $\color{purple}\mathbf{b}_{m_2}^{(\cdot)}$), and still achieve accuracies that rival full-model fine-tuning. 

\section{Experiments and Results}

\noindent\textbf{Datasets.}\quad
We evaluate BitFit on the GLUE\footnote{Appendix \S\ref{app:glue} lists the tasks and evaluation metrics.} benchmark \cite{glue}. Consistent with previous work \cite{DBLP:journals/corr/abs-1902-00751,diffp} we exclude the WNLI task, on which BERT models do not outperform the majority baseline.\\

\noindent\textbf{Models and Optimization.}
We use the publicly available pre-trained BERT\textsubscript{BASE}, BERT\textsubscript{LARGE} \cite{BERT} and RoBERTa\textsubscript{BASE} \cite{ROBERTA} models, using the HuggingFace \cite{huggingface} interface and implementation. Appendix \S\ref{app:opt} lists optimization details. \\

\noindent\textbf{Comparison to Diff-Pruning and Adapters (Table \ref{table:main})}\quad
In the first experiment, we compare BitFit to Diff-Pruning method and Adapters method, when using a fewer number of parameters. Table \ref{table:main} reports the dev-set and test-set performance compared to the Diff-Pruning and Adapters numbers reported by \citet{diffp} and \citet{DBLP:journals/corr/abs-1902-00751} (respectively). This experiment used the BERT\textsubscript{LARGE} model.

On validation set, BitFit outperforms Diff-Pruning on 4 out of 9 tasks, while using 6x fewer trainable parameters\footnote{QNLI results are not directly comparable, as the GLUE benchmark updated the test set since then.}. As for test-set results, two clear wins compared to Diff-Pruning and 4 clear wins compared to Adapters while using 45x fewer trainable parameters.\\

\noindent\textbf{Different Base-models (Table \ref{table:models})}\quad We repeat the BERT\textsubscript{LARGE} results on different base-models (the smaller BERT\textsubscript{BASE} and the better performing RoBERTa\textsubscript{BASE}). The results in Table \ref{table:models} show that the trends remain consistent.

\begin{table*}[h!]
\centering
\scalebox{0.71}{
\begin{tabular}{llccccccccccc}
\hline \textbf{} & \textbf{Method} & \textbf{\%Param} & \textbf{QNLI} & \textbf{SST-2} & \textbf{MNLI\textsubscript{m}} & \textbf{MNLI\textsubscript{mm}} & \textbf{CoLA} & \textbf{MRPC} & \textbf{STS-B} & \textbf{RTE} & \textbf{QQP} & \textbf{Avg.} \\ \hline
 BB & Full-FT & 100\% & \textbf{90.7$\pm$0.2} & 92.0$\pm$0.4 & \textbf{83.5$\pm$0.1} & \textbf{83.7$\pm$0.3} & 56.4$\pm$0.9 & 89.0$\pm$1.0 & 88.9$\pm$0.7 & 70.5$\pm$0.6 & \textbf{87.1$\pm$0.1} & 82.3 \\
 BB & BitFit & 0.09\% & 90.2$\pm$0.2 & \textbf{92.1$\pm$0.3} & 81.4$\pm$0.2 & 82.2$\pm$0.2 & \textbf{58.8$\pm$0.5} & \textbf{90.4$\pm$0.5} & \textbf{89.2$\pm$0.2} & \textbf{72.3$\pm$0.9} & 84.0$\pm$0.2 & \textbf{82.4} \\
 \hline
 BL & Full-FT & 100\% & \textbf{91.7$\pm$0.1} & \textbf{93.4$\pm$0.2} & \textbf{85.5$\pm$0.4} & \textbf{85.7$\pm$0.4} & 62.2$\pm$1.2& 90.7$\pm$0.3 & 90.0$\pm$0.4 & 71.9$\pm$1.3 & \textbf{87.5$\pm$0.4} & 84.1 \\
 BL & BitFit & 0.08\% & 91.4$\pm$2.4 & 93.2$\pm$0.4 & 84.4$\pm$0.2 & 84.8$\pm$0.1 & \textbf{63.6$\pm$0.7} & \textbf{91.7$\pm$0.5} & \textbf{90.3$\pm$0.1} & \textbf{73.2$\pm$3.7} & 85.4$\pm$0.1 & \textbf{84.2} \\
 \hline
 Ro & Full-FT & 100\% & \textbf{92.3$\pm$0.2} & \textbf{94.2$\pm$0.4} & \textbf{86.4$\pm$0.3} & \textbf{86.9$\pm$0.3} & 61.1$\pm$0.8 & \textbf{92.5$\pm$0.4} & 90.6$\pm$0.2 & 77.4$\pm$1.0 & \textbf{88.0$\pm$0.2} & \textbf{85.3} \\
 Ro & BitFit & 0.09\% & 91.3$\pm$0.2 & 93.7$\pm$0.1 & 84.8$\pm$0.1 & 85.2$\pm$0.2 & \textbf{61.8$\pm$1.3} & 92.0$\pm$0.4 & \textbf{90.8$\pm$0.3} & \textbf{77.8$\pm$1.7} & 84.5$\pm$0.2 & 84.6 \\
 
\hline
\end{tabular}}
\caption{\label{table:models} Dev-set results for different base models. \textbf{BB}: BERT\textsubscript{BASE}. \textbf{BL}: BERT\textsubscript{LARGE}. \textbf{Ro}: RoBERTa\textsubscript{BASE}. }
\end{table*}

\begin{figure}[t]
\centering
\includegraphics[width=0.9\linewidth]{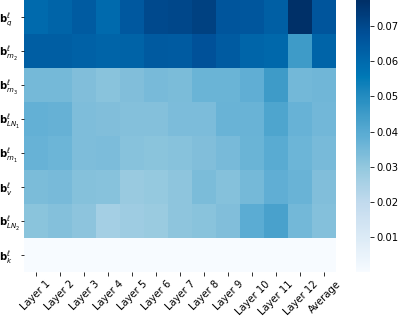}
\caption{Change in bias components (RTE task).} 
\label{fig:heatmap}
\end{figure}

\begin{table*}[h!]
\centering
\scalebox{0.71}{
\begin{tabular}{lccccccccccc}
\hline \textbf{} & \textbf{\% Param} & \textbf{QNLI} & \textbf{SST-2} & \textbf{MNLI\textsubscript{m}} & \textbf{MNLI\textsubscript{mm}} & \textbf{CoLA} & \textbf{MRPC} & \textbf{STS-B} & \textbf{RTE} & \textbf{QQP} & \textbf{Avg.}\\ \hline
 Full-FT & 100\% & \textbf{90.7$\pm$0.2} & 92.0$\pm$0.4 & \textbf{83.5$\pm$0.1} & \textbf{83.7$\pm$0.3} & 56.4$\pm$0.9 & 89.0$\pm$1.0 & 88.9$\pm$0.7 & 70.5$\pm$0.6 & \textbf{87.1$\pm$0.1 } & 82.3\\
BitFit & 0.09\% & 90.2$\pm$0.2 & \textbf{92.1$\pm$0.3} & 81.4$\pm$0.2 & 82.2$\pm$0.2 & \textbf{58.8$\pm$0.5} & \textbf{90.4$\pm$0.5} & \textbf{89.2$\pm$0.2} & \textbf{72.3$\pm$0.9} & 84.0$\pm$0.2  & \textbf{82.4}\\
$\color{purple}\mathbf{b}_{m2},\mathbf{b}_q$ & 0.04\% & 89.4$\pm$0.1 & 91.2$\pm$0.2 & 80.4$\pm$0.2 & 81.5$\pm$0.2 & 57.4$\pm$0.8 & 89.0$\pm$0.2 & 88.4$\pm$0.1 & 68.6$\pm$0.6 & 83.7$\pm$0.2  & 81.1\\
$\color{purple}\mathbf{b}_{m2}$ & 0.03\% & 88.9$\pm$0.1 & 91.1$\pm$0.3 & 79.9$\pm$0.3 & 80.7$\pm$0.2 & 54.9$\pm$0.9 & 87.9$\pm$0.6 & 88.2$\pm$0.1 & 66.8$\pm$0.6 & 82.1$\pm$0.4 & 80.0 \\
$\color{purple}\mathbf{b}_{q}$ & 0.01\% & 86.8$\pm$0.1 & 89.6$\pm$0.2 & 74.4$\pm$0.3 & 75.7$\pm$0.2 & 49.1$\pm$1.5 & 84.4$\pm$0.2 & 85.6$\pm$0.1 & 61.4$\pm$1.1 & 80.6$\pm$0.4 & 76.6 \\
Frozen & 0.0\% & 68.7$\pm$0.3 & 81.7$\pm$0.1 & 42.4$\pm$0.1 & 43.8$\pm$0.1 & 31.9$\pm$1.1 & 81.1$\pm$0.1 & 71.4$\pm$0.1 & 56.9$\pm$0.4 & 62.4$\pm$0.2 & 62.1 \\
rand uniform & 0.09\% & 87.8$\pm$0.3 & 90.5$\pm$0.3 & 78.3$\pm$0.3 & 78.8$\pm$0.2 & 54.1$\pm$1.0 & 84.3$\pm$0.3 & 87.2$\pm$0.4 & 62.9$\pm$0.9 & 82.4$\pm$0.3 & 78.5 \\
rand row/col & 0.09\% & 88.4$\pm$0.2 & 91.0$\pm$0.3 & 79.4$\pm$0.3 & 80.1$\pm$0.3 & 53.4$\pm$0.6 & 88.0$\pm$0.7 & 87.9$\pm$0.2 & 65.1$\pm$0.7 & 82.3$\pm$0.2 & 79.5 \\
\hline

\end{tabular}}
\caption{\label{table:subbias}
Fine-tuning using a subset of the bias parameters. Reported results are for the BERT\textsubscript{BASE} model. }
\end{table*}

\paragraph{Are bias parameters special?} Are the bias parameters special, or will any random subset do? We randomly sampled the same amount of parameters as in BitFit from the entire model, and fine-tuned only them (``rand uniform'' line in Table \ref{table:subbias}). The results are substantially worse across all tasks; similar patterns are observed when the random parameters are sampled as complete rows/columns in the parameter matrices (``rand row/col'' line in Table \ref{table:subbias}).

\paragraph{Fewer bias parameters (Table \ref{table:subbias})}

Can we fine-tune on only a subset of the bias-parameter?

We define the amount of change in a bias vector $\mathbf{b}$ to be $\frac{1}{\mathbf{dim(b)}}\norm{\mathbf{b}_0 - \mathbf{b}_F}_1$, that is, the average absolute change, across its dimensions, between the initial LM values $\mathbf{b}_0$ and its fine-tuned values $\mathbf{b}_F$. Figure \ref{fig:heatmap} shows the change per bias term and layer, for the RTE task (other tasks look very similar, see Appendix \S\ref{app:heatmaps}). The `key' bias $\color{purple}\textbf{b}_{k}$ has zero change, consistent with the theoretical observation in \citet{DBLP:journals/corr/abs-2006-16362}. In contrast,
$\color{purple}\textbf{b}_q$, the bias of the queries, and $\color{purple}\textbf{b}_{m2}$, the bias of the intermediate MLP layers (which take the input from 768-dims to 3072), change the most. Table \ref{table:subbias} reports dev-set results when fine-tuning only the
$\color{purple}\textbf{b}_q^{(\cdot)}$ and $\color{purple}\textbf{b}_{m2}^{(\cdot)}$ bias terms, for the BERT\textsubscript{BASE} model. Results are only marginally lower than when tuning all bias parameters. Tuning either $\color{purple}\textbf{b}_q^{(\cdot)}$ or $\color{purple}\textbf{b}_{m2}^{(\cdot)}$ alone yields substantially worse results, indicating both bias types are essential. As expected, using a frozen BERT\textsubscript{BASE} model yields much worse results. 

\begin{figure}[t!]
\centering
\includegraphics[scale=0.55]{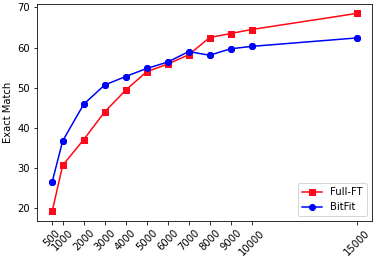}
\caption{Comparison of BitFit and Full-FT with BERT\textsubscript{BASE} exact match score on SQuAD validation set.}
\vspace{-1.2em}
\label{fig:squad-em}
\end{figure}

\paragraph{Generalization gap.}
While in most cases full fine-tuning reaches nearly 100\% train accuracy, we find that the generalization gap \cite{shalev2014understanding}---the difference between training error and test error---is substantially smaller for the BitFit models.
\paragraph{Token-level tasks.} The GLUE tasks are all sentence level. We also experimented with token-level PTB POS-tagging. Full-FT results for BERT\textsubscript{BASE}, BERT\textsubscript{LARGE} and RoBERTa\textsubscript{BASE} are 97.2, 97.4, 97.2, while BitFit results are 97.2, 97.4, 97.1.
\paragraph{Size of training data.} The GLUE results suggest a reverse correlation between BitFit ability to reach Full-FT performance, and training set size. To test this (and to validate another token-level task), we train on increasing-sized subsets of SQuAD v1.0 \citet{squad_v1}. The results on Figure \ref{fig:squad-em} show a clear trend: BitFit dominates over Full-FT in the smaller-data regime, while the trend is reversed when more training data is available. We conclude that BitFit is a worthwhile targetted fine-tuning method in small-to-medium data regimes.

\vspace{-5pt}
\section{Related Work}
\vspace{-5pt}

The problem of identifying the minimal set of parameters that need to be fine-tuned to achieve good performance in end-tasks relates both to practical questions of model compression, and also to more fundamental question on the nature of the pre-training and finetuning process, the ``linguistic knowledge`` induced by each of them, and the extent to which it generalizes to different tasks. 

\noindent\textbf{Over-parameterization.}\quad
 Large LM models were shown to be \emph{over-parameterized}: they contain more parameters than needed in inference \citep{compression-orig, DBLP:journals/corr/HintonVD15, DBLP:conf/iclr/UrbanGKAWMPRC17, DBLP:journals/tnn/Karnin90, DBLP:journals/tnn/Reed93, DBLP:journals/cejcs/AugastaK13, DBLP:conf/interspeech/LiuZW14, han2015learning, DBLP:conf/iclr/MolchanovTKAK17}. \citet{DBLP:journals/corr/abs-2002-08307} have demonstrated that overparmeterization can be exploited in finetuning: pruned network perform well in transfer setting. We work in a complementary setting, where the entire model is kept, but only some parameters are updated.  The remarkable success of those works have sparked interest the lottery-ticket hypothesis \citep{DBLP:conf/iclr/FrankleC19, DBLP:conf/nips/ChenFC0ZWC20, DBLP:conf/emnlp/PrasannaRR20}: the conjecture that large models are needed in pretraining only to induce (in high probability) the existing of sub-networks initialized with the correct inductive bias for learning, and the findings that those sparse networks often transfer well to different tasks. 

\noindent\textbf{Bias terms.}\quad
Bias terms and their importance are rarely discussed in the literature\footnote{Indeed, the equations in the paper introducing the Transformer model \cite{DBLP:journals/corr/VaswaniSPUJGKP17} do not include bias terms at all, and their existence in the BERT models might as well be a fortunate mistake.}. \citet{zhao-etal-2020-masking} describe a masking-based fine-tuning method, and explicitly mention \emph{ignoring} the bias terms, as handling them ``did not observe a positive effect on performance''.\\
\indent An exception is the work of \citet{DBLP:conf/icml/WangZB19} who analyzed bias terms from the perspective of attribution method. They demonstrate that the last layer bias values are responsible for the predicted class, and propose a way to back-propagate their importance. \citet{DBLP:conf/acl/MichelN18} finetuned the biases of the output softmax in an NMT systems, to personalize the output vocabulary, and \citet{DBLP:journals/corr/abs-2003-00152}  have demonstrated that randomly-initialized CNNs achieve reasonable accuracy after training the batch-norm layers alone. Finally, and closest to our work, \citet{tinytl} demonstrate that bias-only fine-tuning similar to ours is effective also for adaptation of pre-trained computer vision models. Our work empirically shows the importance and power of the bias parameters to substantially change the networks' behavior, calling for further analysis and attention on the bias terms.
 
\section{Conclusions}

We propose BitFit, a novel method for localized, fast fine-tuning of pre-trained transformers for end-tasks. The method focuses the finetuning on a specific fraction of the model parameters---the biases---and maintains good performance in all GLUE tasks we evaluated on. The focus on modifying a small group of parameters eases deployment, as the vast majority of the parameters of the model are shared between various NLP tasks. It also allows for efficient hardware implementations that hard-wire most of the network computation with the pre-trained weights, while only allowing few changeable parts for inference time.

Besides its empirical utility, the remarkable effectiveness of bias-only fine-tuning raises intriguing questions on the fine-tuning dynamics of pre-trained transformers, and the relation between the bias terms and transfer between LM and new tasks. 

\section*{Acknowledgments}
This project has received funding from the European Research Council (ERC) under the European Union's Horizon 2020 research and innovation programme, grant agreement No. 802774 (iEXTRACT).

\bibliography{main}
\bibliographystyle{acl_natbib}

\clearpage
\appendix
\section{Appendices}
\label{sec:appendix}

\subsection{Layer naming}
\label{app:names}
For convenience, we relate the notation used in the paper with the names of the corresponding parameters in the popular HuggingFace \cite{huggingface} implementation.

\begin{table}[h!]
\centering
\scalebox{0.93}{
\begin{tabular}{ll}
\hline 
\textbf{HuggingFace Parameter Name} & \textbf{BitFit notation} \\
\hline
attention.self.query.bias & \textcolor{purple}{$\mathbf{b}_{q}$}  \\
attention.self.key.bias & \textcolor{purple}{$\mathbf{b}_{k}$}  \\
attention.self.value.bias & \textcolor{purple}{$\mathbf{b}_{v}$} \\
attention.output.dense.bias & \textcolor{purple}{$\mathbf{b}_{m_1}$} \\
attention.output.LayerNorm.bias & \textcolor{purple}{$\mathbf{b}_{LN_1}$} \\
intermediate.dense.bias & \textcolor{purple}{$\mathbf{b}_{m_2}$} \\
output.dense.bias & \textcolor{purple}{$\mathbf{b}_{m_3}$} \\
output.LayerNorm.bias & \textcolor{purple}{$\mathbf{b}_{LN_2}$} \\
\hline

\end{tabular}}
\caption{\label{table:bias_layer_name_to_symbol} Mapping the HuggingFace's BertLayer bias parameters names to BitFit paper bias notation.}
\end{table}

\subsection{Training Details}
\label{app:opt}
To perform classification with BERT, we follow the approach of \citet{BERT}, and attach a linear layer to the contextual embedding of the \texttt{[CLS]} token to predict the label. The GLUE tasks are fed into BERT using the standard procedures.\\
We optimize using AdamW \cite{adamw}, with batch sizes of 16. For full fine-tuning, we used initial learning rates in \{1e-5, 2e-5, 3e-5, 5e-5\}, and for the bias-only experiments we used initial learning rates in \{1e-4, 4e-4, 7e-4, 1e-3\} as the smaller rates took a very long time to converge on some of the tasks. With the larger learning rates, the bias-only fine-tuning converged in 8 or fewer epochs for most tasks, and up to 20 epochs on the others. We did not perform hyperparameter optimization beyond the minimal search over 4 learning rates. In each evaluation we report X$\pm$Y where X is the average result for training 5 models with 5 different random seeds, Y is the standard deviation.\\
To perform classification with RoBERTa\textsubscript{BASE}, we follow the above details but without hyperparameter search over the learning rates, for bias-only fine-tuning we used 1e-4 as learning rate and for full fine-tuning we used 1e-5 as learning rate.\\
As \citet{bertstability} show, fine-tuning BERT\textsubscript{LARGE} and RoBERTa\textsubscript{BASE} is a unstable due to vanishing gradients. BitFit allows for the usage of bigger learning rates, and overall the optimization process is much more stable, when compared with a full fine-tuning.

\subsection{GLUE Benchmark}
\label{app:glue}
We provide information on the GLUE tasks we evaluated on, as well as on the evaluation metrics. 
We test our approach on the following subset of the GLUE \cite{glue} tasks: The Corpus of Linguistic Acceptability (CoLA; \citet{cola}),
The Stanford Sentiment Treebank (SST-2; \citet{sst2}), The Microsoft Research Paraphrase Corpus (MRPC; \citet{mrpc}), The Quora Question Pairs (QQP; \citet{qqp}), The Semantic Textual Similarity Benchmark (STS-B; \citet{stsb}), The Multi-Genre Natural Language Inference Corpus (MNLI; \citet{mnli}), The Stanford Question Answering Dataset (QNLI; \citet{qnli}) and The Recognizing Textual Entailment (RTE; \citet{rte}).

The metrics that we used to evaluate GLUE Benchmark are in Table \ref{table:glue_metrics}. Learning rate configurations for best performing models are in Table \ref{table:glue_hyperparams}. For all the experiments we used the common train:dev:test partition of GLUE.

\begin{table}[ht]
\centering
\scalebox{0.9}{
\begin{tabular}{ll}
\hline 
\textbf{Task Name} & \textbf{Metric} \\
\hline
QNLI & acc.  \\
SST-2 & acc.  \\
MNLI & matched acc./mismatched acc. \\
CoLA & Matthews corr. \\
MRPC & F1 \\
STS-B & Spearman corr. \\
RTE & acc. \\
QQP & F1 \\
\hline

\end{tabular}}
\caption{\label{table:glue_metrics} Metrics that we use to evaluate GLUE Benchmark.}
\end{table}

\begin{table}[ht]
\centering
\scalebox{0.95}{
\begin{tabular}{lcc}
\hline 
\textbf{Task Name} & \textbf{BERT\textsubscript{BASE}} & \textbf{BERT\textsubscript{LARGE}}\\
\hline
QNLI & 1e-4 & 7e-4 \\
SST-2 & 4e-4 & 4e-4 \\
MNLI & 1e-4 & 1e-4 \\
CoLA & 7e-4 & 4e-4 \\
MRPC & 7e-4 & 1e-3 \\
STS-B & 1e-4 & 1e-4 \\
RTE & 1e-3 & 4e-4 \\
QQP & 4e-4 & 4e-4 \\
\hline

\end{tabular}}
\caption{\label{table:glue_hyperparams} Learning rate configurations for best performing models.}
\end{table}

\newpage
\subsection{Amount of change in bias terms}
\label{app:heatmaps}
\begin{figure}[h!]
\centering
\includegraphics[width=\linewidth]{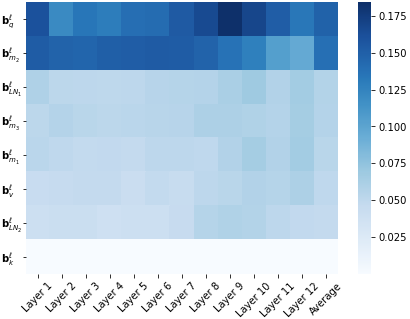}
\caption{Change in bias components (CoLA task).} 
\label{fig:cola_heatmap}
\end{figure}

\begin{figure}[h!]
\centering
\includegraphics[width=\linewidth]{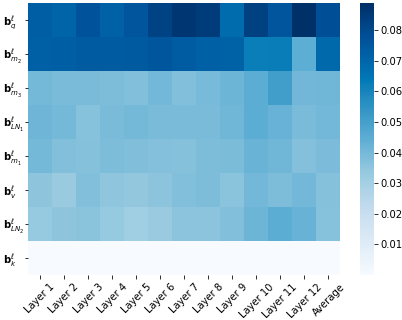}
\caption{Change in bias components (MRPC task).} 
\label{fig:mrpc_heatmap}
\end{figure}

\begin{figure}[h!]
\centering
\includegraphics[width=\linewidth]{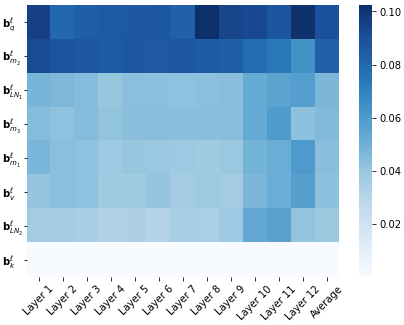}
\caption{Change in bias components (STS-B task).} 
\label{fig:stsb_heatmap}
\end{figure}

\newpage



\subsection{SQuAD F1 Results}

\begin{figure}[!htb]
\centering
\includegraphics[scale=0.55]{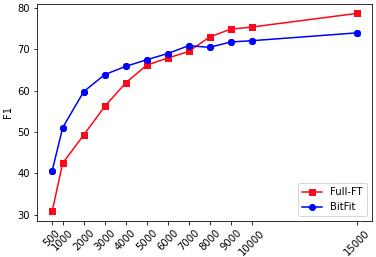}
\caption{Comparison of BitFit and Full-FT with BERT\textsubscript{BASE} F1 score on SQuAD validation set.}
\label{fig:squad-f1}
\end{figure}

\end{document}